%% file: main.tex
\documentclass{article}

\usepackage{amsthm}
\usepackage{amsmath}
\usepackage{multirow}
\usepackage{amsfonts}
\usepackage{graphics}
\usepackage{mathtools}
\usepackage{microtype}
\usepackage{graphicx}
\usepackage{subfigure}
\usepackage{booktabs} 

\usepackage{hyperref}

\usepackage[accepted]{mlsys2024}

\mlsystitlerunning{Quality at the Tail of Machine Learning Inference}

\begin{document}

\twocolumn[
\mlsystitle{Quality at the Tail of Machine Learning Inference}



\mlsyssetsymbol{equal}{*}

\begin{mlsysauthorlist}
\mlsysauthor{Zhengxin Yang}{ictcas,ucas}
\mlsysauthor{Wanling Gao}{ictcas,ucas}
\mlsysauthor{Chunjie Luo}{ictcas,ucas}
\mlsysauthor{Lei Wang}{ictcas,ucas}
\mlsysauthor{Fei Tang}{ictcas,ucas}
\mlsysauthor{Xu Wen}{ictcas,ucas}
\mlsysauthor{Jianfeng Zhan}{ictcas,ucas}
\end{mlsysauthorlist}

\mlsysaffiliation{ictcas}{Institute of Computing Technology, Chinese Academy of Sciences}
\mlsysaffiliation{ucas}{University of Chinese Academy of Sciences}

\mlsyscorrespondingauthor{Jianfeng Zhan}{zhanjianfeng@ict.ac.cn}

\mlsyskeywords{Machine Learning, MLSys}

\vskip 0.3in

\begin{abstract}
Machine learning inference should be subject to stringent inference time constraints while ensuring high inference quality, especially in safety-critical (e.g., autonomous driving) and mission-critical (e.g., emotion recognition) contexts.
Neglecting either aspect can lead to severe consequences, such as loss of life and property damage.
Many studies lack a comprehensive consideration of these metrics, leading to incomplete or misleading evaluations.
The study unveils a counterintuitive revelation: deep learning inference quality exhibits fluctuations due to inference time.
To depict this phenomenon, the authors coin a new term, ``tail quality,'' providing a more comprehensive evaluation, and overcoming conventional metric limitations.
Moreover, the research proposes an initial evaluation framework to analyze factors affecting quality fluctuations, facilitating the prediction of the potential distribution of inference quality.
The effectiveness of the evaluation framework is validated through experiments conducted on deep learning models for three different tasks across four systems.
\end{abstract}
]



\printAffiliationsAndNotice{}  

\input{sections/introduction}


\input{sections/quality}

\input{sections/experiments}

\input{sections/related}

\input{sections/conclusion}

\bibliography{reference}
\bibliographystyle{mlsys2024}



\end{document}

%% file: sections/introduction.tex
\section{Introduction}\label{introduction}

Deep learning is catalyzing rapid advancements across a wide range of domains, encompassing areas such as natural language processing~\cite{vaswaniAttentionAllYou2017,devlinBERTPretrainingDeep2019,zhengJudgingLLMasajudgeMTBench2023} and computer vision~\cite{goodfellowGenerativeAdversarialNets2014,heDeepResidualLearning2016,carionEndtoEndObjectDetection2020,dosovitskiyImageWorth16x162023}.
This surge is paving the way for an escalating deployment of deep learning models and systems in cutting-edge real-world applications such as autonomous driving~\cite{chenDeepDrivingLearningAffordance2015,xuEndtoEndLearningDriving2017,liDeepLearningApproaches2020} and smart healthcare~\cite{rajpurkarDeepLearningChest2018,mckinneyAddendumInternationalEvaluation2020}.
In the face of this expansion across diverse fields, the execution of benchmarking and evaluation is essential to ensure the proper and effective development of these models and systems.
Furthermore, within real-world applications, it is paramount to consider the metrics of inference quality and inference time carefully and comprehensively when establishing benchmarking and evaluation methodologies.
These metrics reflect the capabilities of deep learning and significantly influence the user experience.

Unfortunately, many current studies tend to focus on evaluating only one aspect while neglecting the other and fail to consider the evaluation in real-world applications, especially critical\footnote{Based on \citet{sommervilleSoftwareEngineering2011}, ``critical systems'' are systems where ``system failure may result in injury to people, damage to the environment, or extensive economic losses.'' Examples include autonomous driving (safety-critical), business negotiation (business-critical), and navigational system for a spacecraft (mission-critical)~\cite{gaoHighFusionComputers2022}.} tasks.
One significant reason for this issue is that it is commonly believed that once the neural network model is trained and its parameters are fixed, the inference quality of deep learning on the same data will never change regardless of variations in the deployment environment.
However, as this paper is about to reveal, a counterintuitive phenomenon in the practical deployment and application of deep learning is captured, in which fluctuations in inference quality can occur due to variations in inference time, even when the inference inputs remain unchanged.
Specifically, for safety-critical tasks like autonomic driving,  ensuring high-quality inference while adhering to stringent inference time requirements is of utmost importance.
For example, given that humans typically need about 390 to 600 milliseconds to respond to hazards~\cite{wolfeRapidholisticperception2020,wolfeEffectstemporalspatiotemporal2021}, the capability of object detection systems in autonomous driving must surpass human levels\cite{turayPerformingImageClassification2022} within mere tens to hundreds of milliseconds, to reduce the likelihood of accidents.
For a specific critical task object detection model, even if the inference time exceeds the specific time constraint by just a few tens of milliseconds, it may have already traveled several meters and caused severe consequences such as loss of life and property damage.

To better illustrate this phenomenon, the authors propose a new term \emph{``tail quality.''}
Specifically, tail quality @$x\%$=$t$ identifies inference quality when the inference time threshold is set as $x\%$ tail latency $t$ (e.g. $99.9\%$ tail latency = $10$ms (milliseconds) as the threshold).
For a hard real-time system, it's essential to ensure that the tail quality @$100\%$=$t$ remains high even when $t$ is minimal, guaranteeing the reliability of the system.
If the acceptable threshold is $10$ms, this means the tail quality @$100\%$=$10$ms must be sufficiently high.

\begin{figure}[!htbp]
\centering
\includegraphics[width=0.43\textwidth]{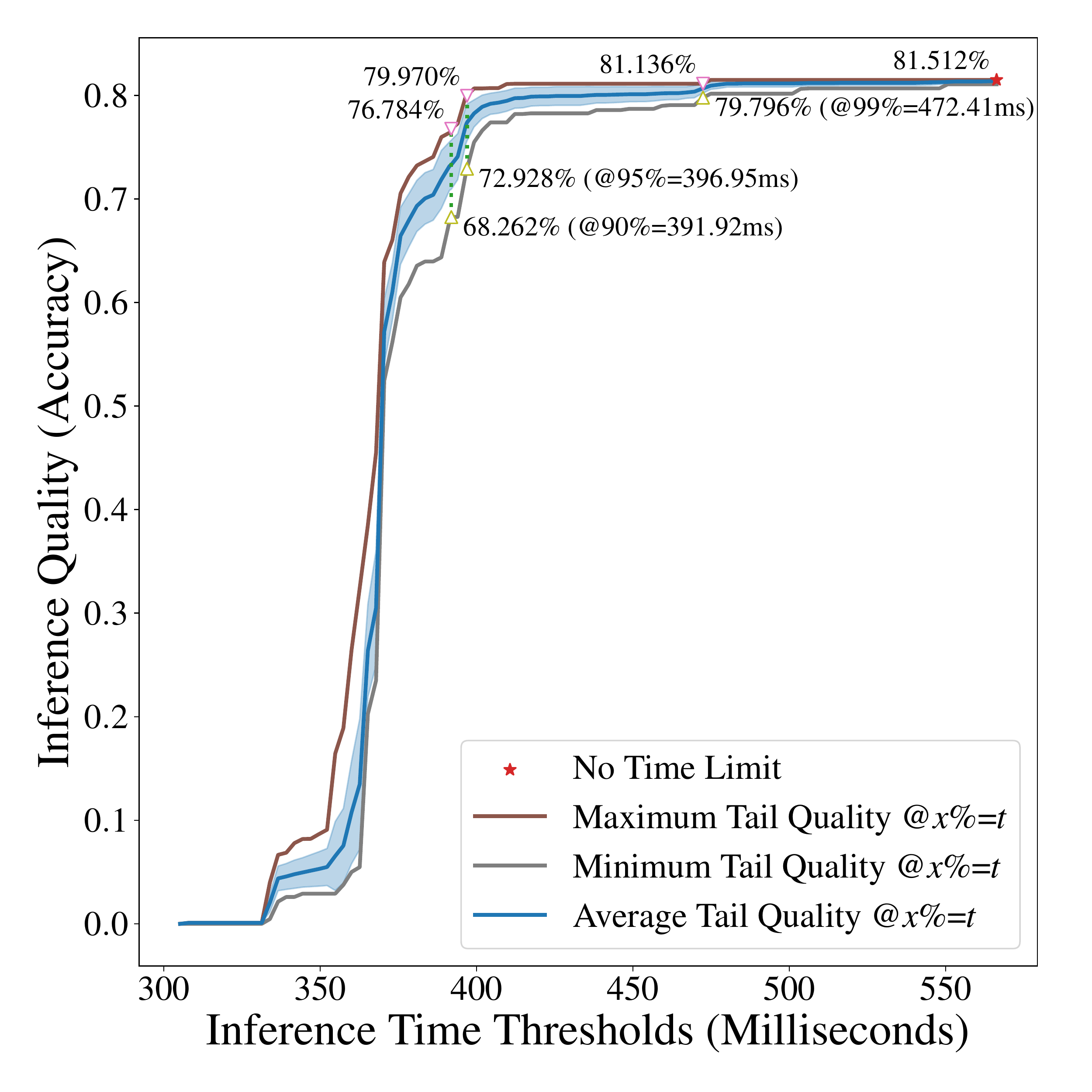}
\vskip -0.1in
\caption{
Quality Fluctuations of an Image Classification Model Vision Transformer (ViT) on Tesla P100. The triangular symbols from left to right represent the maximum and minimum values of inference quality, using the 90th, 95th, and 99th percentiles of tail latency as the inference time thresholds.
}
\label{fig:tail_quality}
\end{figure}
Figure~\ref{fig:tail_quality} provides a more intuitive depiction of the phenomenon.
As observed, compared to the original inference quality without time constraints, where tail quality @$100\%$=$t$ is $81.512\%$ accuracy, the worst-case tail quality @$99$\%=$472$ms dropped by approximately 1.7 percentage points.
More seriously, the worst-case tail quality @$90$\%=$392$ms plummeted by around 13 percentage points.
As the threshold becomes stricter, moving from $400$ms to $300$ms, the model's inference quality exhibited significant fluctuations and declined dramatically, eventually reaching an accuracy of $0$\%.

However, such threshold settings are stringent in safety-critical tasks; for instance, tasks like autonomous driving need to achieve high quality when inference times are within a few tens of milliseconds, e.g., the tail quality @$100$\%=$10$ms should be higher than $99.9$\% accuracy, or irreversible consequences may occur.
This highlights the severity of the ``tail quality'' phenomenon.
``Tail quality'' could have profound implications in real-world scenarios, particularly in contexts like driving, where decisions are constantly made based on rapidly changing traffic conditions.
Considering the enormous number of vehicles on the roads daily, if control were to be handed over to a deep learning model, even if the probability of encountering a ``tail quality'' event (i.e., inference failure) is low, within such a vast population, any decision-making error arising from such an occurrence could lead to loss of life or property, which is unacceptable.

Many studies have predominantly used individual quality metrics such as \emph{accuracy}~\cite{heDeepResidualLearning2016,dosovitskiyImageWorth16x162023}, \emph{average precision}~\cite{everinghamPascalVisualObject2010,linMicrosoftCOCOCommon2014,carionEndtoEndObjectDetection2020}, or individual inference time metrics like \emph{tail latency}~\cite{deanTailScale2013,reddiMLPerfInferenceBenchmark2020,gaoAIBenchIndustryStandard2019,gaoAIBenchScalableComprehensive2019} to characterize the performance of deep learning.
What sets apart the concept of \emph{``tail quality''} is that it allows for a more intuitive depiction of the impact of inference time on inference quality and thus provides insights into the extent of potential consequences caused by inference failures.
To the best of our knowledge, this paper is the first to present this discovery.

Due to the intricacies of benchmarking in computer science, deep learning software and hardware systems, along with the models that serve as their workloads, are entangled and mutually influential~\cite{reddiMLPerfInferenceBenchmark2020,zhanBenchCouncilViewBenchmarking2022}.
Conducting a systematic analysis of the causes behind fluctuations in deep learning inference quality has been challenging.
Additionally, with the aim of promoting extensive research in this direction, this research proposes an initial evaluation framework to analyze various factors that affect the fluctuation of inference quality.
It systematically analyzes the effects of the entire deployment and application environment of deep learning by disassembling it into several components, such as software and hardware systems, models, and data.
Furthermore, statistical methods intuitively depict the approximate distribution of inference time and inference quality under the influence of different components, enabling the prediction of the ``tail quality'' phenomenon before deploying deep learning applications.

Experiments are conducted on frequently used deep learning models for three different tasks across four systems and two deep learning frameworks.
On the one hand, through experiments, it has been further validated that the superiority of adopting the proposed novel term \emph{``tail quality''} can provide a more intuitive characterization of the severe consequences of instability in deep learning models and systems in real-world applications compared to relying solely on inference quality or inference time as evaluation metrics.
On the other hand, the experiments confirmed the effectiveness of this evaluation framework in predicting ``tail quality.''
During the testing phase, the evaluation framework achieved an average squared root of Jensen-Shannon Divergence (rJSD) value of 0.051 for the predicted probability distribution of inference times.
In predicting the worst-case tail quality @$99$\%=$t$,  @$95$\%=$t$, and @$90$\%=$t$, it exhibited an average discrepancy level of -0.07, indicating its ability to reasonably forecast the worst-case tail quality.
In terms of the framework performance, MLPerf Inference~\cite{reddiMLPerfInferenceBenchmark2020} a minimum of 262,742 inferences to determine the 99th percentile Tail Latency.
Furthermore, the proposed evaluation framework only requires an average of only 62.26\% of the inference count needed by MLPerf Inference across all experiments.

As the first work in the novel research direction, the authors urge caution in dealing with the discovered phenomenon of ``tail quality'' and call for establishing innovative methodologies and tools based on the proposed evaluation framework to tackle the challenge above.

%% file: sections/quality.tex
\section{Tail Quality and Evaluation Framework}\label{quality}
In order to proactively anticipate and prevent the occurrence of ``tail quality,'' this section will begin by providing its definition.
This definition will serve as the foundation for the systematic construction of tail quality evaluation framework, enabling a scientifically grounded approach.
Subsequently, an evaluation framework for ``tail quality'' will be introduced.
This framework enables the prediction of potential future instances of tail quality in deep learning models with relatively low computational costs as shown in Section~\ref{experiments}.
Additionally, it facilitates the analysis of potential factors that influence the emergence of tail quality.

\subsection{Definition of Tail Quality}
During deep learning inference phase, given a dataset $ D=\{x_{i}, y_{i}\}_{i=1}^{n} $ containing $ n=\left\lvert D \right\rvert $ instances, model $ M $ reads input instances $ x_{i} $ from the dataset $ D $, generating a set of inference results $ Y^{'}=\{y_{i}^{'}\}_{i=1}^{n} $ where $y_{i}^{'}=M(x_{i})$.
By comparing the difference between inference results $ Y^{'} $ and the ground truth answers $ Y=\{y_{i}\}_{i=1}^{n} $, inference quality $ Q $ of model $ M $ over the entire dataset $ D $ can be obtained.
The calculation method of inference quality depends on the specific quality evaluation metric used, such as accuracy, AP, and F-score. Thus, the calculation process can be abstracted by using the metric calculation function $q$, thus $Q$ can be represented as $Q = q(\{M(x_{i})\}_{i=1}^{n}, \{y_{i}\}_{i=1}^{n})$.
It is evident that when calculating inference quality, the inference result of each instance contributes to the overall inference quality $ Q $ accordingly.
The specific contribution of each instance $ x_{i} $ can be abstracted into a contribution function $ c_{i} $.
Therefore, the quality $ Q $ can be further abstracted as $Q = q(\{c_{i}(x_{i}, y_{i}, M)\}_{i=1}^{n})$.
In a specific task, assume the maximum allowable inference time is denoted as $ \theta $, which serves as the inference time threshold.
The inference time taken by the model $ M $ on a particular instance $ x_{i} $ is denoted as $ t_{i}(x_{i}, M) $.
Thus, the following indicator function $ \mathbf{1}_{i} $ can be used to determine the effectiveness of the inference result:
\begin{equation}\label{eq:validity}
    {\mathbf{1}}_{\theta}(x_{i}, M) = {
        \begin{cases}
            1~&~t_{i}(x_{i}, M) \leq \theta~,\\
            0~&~t_{i}(x_{i}, M) > \theta~
        \end{cases}
    } .
\end{equation}
Additionally, the instability of deep learning systems leads to constant fluctuations in model inference time. Thus, to better investigate the impact of inference time on inference quality, it can be assumed that the inference time $ t_{i}(x_{i}, M) $ of model $ M $ on instance $ x_{i} $ follows a certain conditional probability distribution $P(T=t \mid \mathcal{X}=x_{i}, \mathcal{M}=M)$,
where $ T $, $ \mathcal{X} $, and $ \mathcal{M} $ are three random variables representing time, instance, and model, respectively.
$ t_{i}(x_{i}, M) $ is replaced by $ t $ to make the expression more concise.
Note that the specific model is already determined during the deep learning inference phase.

Ultimately, under the constraint of inference time, the overall inference quality $ Q $ can be represented as:
\begin{equation}\label{eq:detailed_inference_quality}
    Q = q(\{c_i(x_{i}, y_{i}, M) \cdot {\mathbf{1}}_{\theta}(x_{i}, M)\}_{i=1}^{n}) .
\end{equation}
Based on the formulas provided above, it becomes evident that once deterministic inference quality algorithm is now reliant on the inference time of each instance, introducing considerable uncertainty in the calculation inference quality due to considering inference time constraints.

Note that the above analysis applies when the input is a batch; the probability distribution and indicator function corresponding to each instance in the batch can be simply replaced by the ones corresponding to the entire batch.

\subsection{Establishment of Evaluation Framework}
This section introduces a novel evaluation framework which models the entire evaluation process from the system, model, data, and other potential components to the inference time and then to the inference quality, based on the definition of ``tail quality'' from the previous subsection.
It allows researchers to adjust the modeling formulas and parameters to guide the establishment of evaluation processes with varying levels of granularity and precision.
This provides the evaluation framework with high flexibility and scalability.

The variation of each component can potentially impact the probability distribution of the deep learning inference time.
Therefore, the probability distribution model that describes and analyzes the various influencing factors that impact the inference time of deep learning can be represented as the conditional probability distribution of inference time $P(T \mid \mathcal{C}^{1}, \mathcal{C}^{2}, \cdots)$,
where $ \mathcal{C}^{j} $ denotes the influencing components such as model $ \mathcal{M} $.
In a practical application of the framework, components that are not considered will not be listed in the formula.
Therefore, to ensure the objectivity of the analysis, the evaluation environment for deep learning inference should remain unchanged except for the listed components.

Building on this inference time distribution model, the inference quality represented by Formula~\ref{eq:detailed_inference_quality} can be rewritten to create the model that describes and analyzes the impact of inference time on inference quality, as shown below:
\begin{equation}\label{eq:frame_quality}
    Q = q(\{c_i(x_{i}, y_{i}, M) \cdot {\mathbf{1}}_{\theta}(x_{i}, M, \{\mathcal{C}^{j}\}_{j=1}^{\infty})\}_{i=1}^{n})
\end{equation}
where the characteristic function $ {\mathbf{1}}_{\theta}(x_{i}, M, \{\mathcal{C}^{j}\}_{j=1}^{\infty}) $ is derived from equation~\ref{eq:validity}.

Therefore, the construction of the evaluation framework mainly consists of two parts: (1) establishing a probability distribution model for inference time, $P(T \mid \mathcal{C}^{1}, \mathcal{C}^{2}, \cdots)$, through which the relationship between individual components and inference time can be determined; (2) constructing a ``tail quality'' calculation model to establish the mapping relationship between inference quality and inference time through this model, as Equation~\ref{eq:frame_quality} represents.

\subsubsection*{Inference Time Probability Model}
In this section, a heuristic algorithm is proposed to estimate the probability distribution of inference times, which can be seen as a basic Monte Carlo simulation. 
As the probability distribution governing the inference time of each instance is unknown, this part proposes a heuristic algorithm to estimate the probability distribution of inference time, in order to construct the inference time probability model.

Algorithm~\ref{alg:time} shows details of the proposed algorithm.
Initially, the deep learning model $ M $ conducted $ r $ rounds of inference for all instances $ x_{i} $ in the dataset $ D $ where $ \left\lvert D \right\rvert = n $, recording the corresponding inference times, and then, by using the method $ {\rm Fit}() $, the probability density functions (PDF) for each instance are fitted based on the initial recorded $ r $ rounds of inference times (line 2-8).
Researchers can opt for an appropriate function-fitting method to model the probability distribution of inference times for each instance.
In this study, the kernel density estimation (KDE) method is chosen to employ.
According to the definition of KDE, the fitted PDF of the instance $ x_{i} $ is as follows:
\begin{equation}\label{eq:kde}
    f_{i}(t) = {\rm {Fit}}(\{t_{i, j}\}_{j=1}^{r}) = {\frac{1}{nh}}\sum_{j=1}^{r}K({\frac{t-t_{i, j}}{h}}) ,
\end{equation}
where $ t_{i, j} $ is the $ j $th round inference time on the instance $ x_{i} $, $ K $ is the kernel function which is chosen as a Gaussian distribution in the paper, and $ h $ is a smoothing parameter.

The algorithm then proceeds to perform deep learning inference on the entire dataset repeatedly until it deems that the PDF for all instances has been sufficiently well-fitted (lines 12-36).
Firstly, during each inference, the inference times on all instances are recorded (line 15).
Then, on the entire dataset, for every $ s $ rounds of inference, the probability distribution models for the inference times of all instances are re-fitted based on the newly recorded $ s $ inference times as well as all the previously recorded inference times (line 16, 17).
The newly fitted model $ F_{i, l+1} $ is then compared with the previously fitted models.

\input{algs/alg_time}
With an increase in the number of inference rounds, the accumulation of more recorded inference times yields additional information about the population of inference times.
This enables a more accurate estimation and prediction of probability distribution.
Therefore, when newly recorded inference times cease to contribute information significantly to the estimation of the probability distribution, indicating that the fitting of the distribution hardly changes, it is considered that a reasonably good estimation of the overall distribution has been achieved.

Based on these considerations, Algorithm~\ref{alg:time} employs a sliding window to store the previous $ w $ fitted distributions.
If the differences between these fitted distributions within the sliding window are sufficiently small, i.e., they are less than the specified tolerance threshold $ \delta $, it is inferred that a sufficiently good estimation of the statistical population can be obtained based on the available inference time records (line 20-25).

The algorithm employs the $ {\rm Check}() $ method to compare differences among different fitting results (line 22).
This method can be specified by researchers as long as it ensures the convergence of the algorithm is not compromised.
In this paper, the $ {\rm Check}() $ method employs Jensen-Shannon (JS) divergence, which is a symmetrized and smoothed version of the Kullback-Leibler (KL) divergence, to quantify the similarity between any two probability distributions.
The definition of the JS divergence is:
\begin{equation}\label{eq:jsd}
\begin{split}
{\rm {JSD}}(P \parallel Q) &= {[D(P \parallel M) + D(Q \parallel M)]}/{2} ,
\end{split}
\end{equation}
where $ P $ and $ Q $ are the probability distributions to be compared, they correspond to $ F_{i, k} $ and $ F_{i, l+1} $ respectively in the context of Algorithm~\ref{alg:time}, and $ M = \frac{1}{2}(P + Q) $ represents the mixture distribution of $ P $ and $ Q $.
$ {\rm {D}}(P \parallel Q) $ denotes the calculation of the KL divergence between $ P $ and $ Q $.

Ultimately, after finishing the estimation of the inference time distribution for all instances, it can be inferred that the fitted probability density functions $ {f_{i}}_{i=1}^{n} $ of each instance and the minimum required number of inference rounds to obtain these density functions.
The computation cost of this algorithm mainly depends on the number of iterations needed to fit PDFs of inference times for all samples.
Consequently, the total number of inferences should be the product of the dataset size and the number of iterations.
Through experiments in Section~\ref{experiments}, it can be observed that the algorithm's fitting convergence rate is quite satisfactory and, in most cases, it outperforms the inference count required by MLPerf Inference~\cite{reddiMLPerfInferenceBenchmark2020}.

\subsubsection*{Tail Quality Calculation Model}
As evident from the previous sections, deriving an analysis model for inference quality directly from the inference time analysis model through Formula~\ref{eq:frame_quality} is quite challenging.
Given the variety of evaluation metrics, it is not feasible to construct quality calculation functions for each metric and to estimate the contribution functions of all instances under specific quality indicators.

Fortunately, the process of estimating the inference time distribution model involves a comprehensive exploration of all potential deep learning inference conditions that may occur.
Hence, given all the recorded inference times, researchers can easily calculate the inference quality for each round of inference at a specific threshold $ \theta $ by utilizing Algorithm~\ref{alg:quality}, which is derived from the Equations~\ref{eq:frame_quality}.

In Algorithm~\ref{alg:quality}, the inference result effectiveness of each instance is tagged by comparing its inference time with the threshold $ \theta $ (line 4-8).
Deep learning inference results that surpass the inference time threshold will be marked as invalid and considered as erroneous outcomes in the overall statistical evaluation of inference quality.
After each round of tagging is completed, the inference quality can be recalculated by using method $ {\rm Evaluate}() $ based on the effectiveness of all instances in the dataset (line 10).
Implementing the $ {\rm Evaluate}() $ method depends on a specific inference quality metric.
For instance, in the case of accuracy evaluation metric, all invalid instances (i.e., instances where $v_{j}=False$) can be considered as errors when assessing correctness against the ground truth.
This means they are not included in the count of correct samples, while other valid instances are counted according to the original calculation method.

\input{algs/alg_quality}
Finally, with the calculated qualities through Algorithm~\ref{alg:quality}, researchers can directly estimate the probability distribution of the inference quality under the specified inference time threshold $ \theta $, which makes the inference quality analysis model more concrete.
However, in practical applications, estimating this distribution model is not always necessary.
The emphasis should lie on particular attributes crucial for real-world applications, such as the worst-case tail quality @$99$\%=$t$ with specific designated inference time constraint.

%% file: algs/alg_time.tex
\begin{algorithm}[!tb]
\caption{Estimation of Inference Time Analysis Model}\label{alg:time}
\begin{algorithmic}[1]

  \STATE{{\bfseries Input:}
    $ M $, $ \{x_{i}\}_{i=1}^{n} $, $ n $, $ r $, $ s $, $ \delta $, and $ w $
  }
  \STATE{{\bfseries Output:}
    $ r $, $ \{f_{i}\}_{i=1}^{n} $, and $ \{\hat{y_{i}}\}_{i=1}^{n} $
  }
  \\
  // Initialization
  \FORALL{ $ i \gets 1 \ldots n $ }
    \FORALL{ $ j \gets 1 \ldots r $ }
      \STATE{ $ t_{i, j} \gets {\rm Timing}(M(x_{i})) $ }
    \ENDFOR
    \STATE{ $ \hat{y_{i}} \gets M(x_{i}) $ }
    \STATE{ $ F_{i, 1} \gets {\rm Fit}(\{t_{i, j}\}_{j=1}^{r}) $ }
  \ENDFOR
  \\

  // Iteration
  \STATE{ $ l \gets 1 $ }
  \REPEAT
    \STATE{ $ r \gets r + l $ }
    \FORALL{ $ i \gets 1 \ldots n $ }
      \STATE{ $ t_{i, j} \gets {\rm Timing}(M(x_{i})) $ }

      \IF{ $ l \mod s = 0 $ {\bf and} $ fit_{i} \neq True $ }
        \STATE{ $ F_{i, l+1} \gets {\rm Fit}(\{t_{i, j}\}_{j=1}^{r}) $ }
        \STATE{ $ f_{i} \gets F_{i, l+1} $ }
        \IF{ $ l+1 > w $ }
          \FORALL{ $ k \gets l+1-w \ldots l $ }
            \STATE{ $ fit_{i} \gets True $ }
            \IF{ $ {\rm Check}(F_{i, k}, F_{i, l+1}) > \delta $ }
              \STATE{ $ fit_{i} \gets False $ }
            \ENDIF
          \ENDFOR
        \ENDIF
      \ENDIF
    
    \ENDFOR
    \STATE{ $ finish \gets True $ }
    \FORALL{ $ i \gets 1 \ldots n $ }
      \IF{ $ fit_{i} \neq True $ }
        \STATE{ $ finish \gets False $ }
      \ENDIF
    \ENDFOR
    \STATE{ $ l \gets l + 1 $ }
  \UNTIL{ $ finish = True $ }
\end{algorithmic}
\end{algorithm}

%% file: algs/alg_quality.tex
\begin{algorithm}[!htbp]
\caption{Calculation of Inference Quality}\label{alg:quality}
\begin{algorithmic}[1]

  \STATE{{\bfseries Input:}
    $ M $, $ r $, $ \theta $, $ \{x_{i}, y_{i}\}_{i=1}^{n} $, $ \{\hat{y}_{i}\}_{i=1}^{n} $, $ \{t_{i, j}\} (i \in [1, n], j \in [1, r]) $
  }
  \STATE{{\bfseries Output:}
    Inference Qualities $ \{q_{i}\}_{i=1}^{r} $,
  }
  \\
  \FORALL{ $ i \gets 1 \ldots r $ }
    \FORALL{ $ j \gets 1 \ldots n $ }
      \IF{ $ t_{j, i} \leq \theta $ }
        \STATE{ $ v_{j} \gets True $ }
      \ELSE
        \STATE{ $ v_{j} \gets False $ }
      \ENDIF
    \ENDFOR
    \STATE{ $ q_{i} \gets {\rm Evaluate}(\{x_{j}, \hat{y}_{j}, v_{j}\}_{j=1}^{n}) $ }
  \ENDFOR
\end{algorithmic}
\end{algorithm}

%% file: sections/experiments.tex
\section{Experimental Analysis and Results}\label{experiments}
In this section, the authors first instantiate the proposed evaluation framework and validate its effectiveness through experiments.
Subsequently, the instantiated evaluation framework is employed to analyze various factors, such as systems and data, that impact inference time and inference quality separately.

\subsection{Instantiation of the Evaluation Framework}
The instantiation of the evaluation framework essentially involves defining the analysis models for inference time and inference quality and selecting appropriate parameters for the estimation Algorithms~\ref{alg:time} and ~\ref{alg:quality}.
For the sake of experiments simplicity, this study considers four influencing factors: hardware systems $\mathcal{S}$, deep learning frameworks $\mathcal{F}$, models $\mathcal{M}$, and data $\mathcal{X}$.
Thus, the instantiation can be represented as $ P(T \mid \mathcal{X}, \mathcal{M}, \mathcal{S}, \mathcal{F}) $.

To define the search space for influencing components, four servers, A, B, C and D are selected with different types of graphics processing units (GPU) as subjects for hardware systems $\mathcal{S}$ investigation in experiments.
Among them, Server A is equipped with 4 identical GeForce RTX 2080 Ti GPUs, facilitating distributed inference for large language models (LLM).
Server B, C, and D is equipped with TITAN V, Tesla P100 and V100, respectively.
For deep learning frameworks $\mathcal{F}$, experiments are conducted under the two most popular frameworks, PyTorch~\cite{paszkePyTorchImperativeStyle2019} and TensorFlow~\cite{abadiTensorFlowsystemlargescale2016}.
Noting that the framework versions are consistent across all servers, and the CUDA version is 11.7 for all except Server C with 11.6.

In order to ensure comprehensiveness while maintaining simplicity in experiments, for models$\mathcal{M}$ and data$\mathcal{X}$ inference components, three widely used models were selected across the Computer Vision (CV) and Natural Language Processing (NLP) domains.
These models include Detection Transformer (DETR)~\cite{carionEndtoEndObjectDetection2020} for object detection, Vision Transformer (ViT)~\cite{dosovitskiyImageWorth16x162023} for image classification, and the large language model Vicuna~\cite{zhengJudgingLLMasajudgeMTBench2023} for dialogue systems (Chatbot).
Corresponding to the selected models, three datasets used to evaluate the model inference quality, namely COCO~\cite{linMicrosoftCOCOCommon2014} val2017, ImageNet~\cite{russakovskyImageNetLargeScale2015} val2012, and MMLU~\cite{hendrycksMeasuringMassiveMultitask2020} dev, were chosen for conducting the experiments.
Table~\ref{tab:task_specs} contains specific information about the data and models.

\input{tables/task_specs}
Another part of the framework instantiation involves selecting hyperparameters for the estimation Algorithm~\ref{alg:time} of the inference time probability model.
Specifically, the initial number $ r $ of inference rounds is set to 30, the size of sliding window $ w $ for storing fitted analysis models is set to 5, and the step size $ s $ for re-fitting is set to 5. The tolerance $ \delta $ for similarity differences between the fitted models is set to 0.2.
It's important to note that the evaluation metric for tolerance difference is the square root of Jensen-Shannon Divergence (rJSD).
JSD ranges from 0 to 1, where lower values indicate more remarkable similarity between two fitted results, with 0 representing identical.

\subsection{Validation of Effectiveness}
\input{tables/verify_time}
To validate the effectiveness of the evaluation framework, the validation process is divided into two stages: the training and testing stages.

\noindent\textbf{In training stage}, inference time probability models are estimated on different servers and deep learning frameworks for each model and its corresponding dataset.
A probability density distribution (PDF) is fitted for each instance based on its inference times.
The fitted PDF should stabilize as the algorithm iterates, indicating that the rJSD between the re-fitted PDF and the existing $ w $ PDFs, which were fitted using a smaller amount of inference time data, should gradually approach 0.
Therefore, rJSD serves as a metric to assess the quality of the fitting results for each instance.
Furthermore, the mean rJSD across all instances in the dataset is employed as an indicator of the overall quality of the algorithm in fitting the entire dataset.

\noindent\textbf{In testing phase}, to assess the generalization performance of PDF of inference times corresponding to each fitted instance by the algorithm, all deep learning models undergo an additional 30 rounds of inference on their respective datasets.
This process generated new inference time data points, which were then used to validate the generalization performance of the probability distributions that are fitted during the training phase.
A better generalization performance indicates that the fitting results encompass sufficient information about the population, capturing a broader range of scenarios that deep learning models might encounter during repeated inferences on the same instance.
Specifically, during the testing phase, probability distribution are fitted to all the inference time data obtained for each instance in the dataset.
Subsequently, the fitted results obtained for each sample during the training phase are individually compared with the corresponding testing phase results, and the rJSD is calculated for each comparison.
The overall generalization performance of the estimated inference time probability model across the whole dataset can be represented by the average rJSD computed for all instances.

The validation of the tail quality calculation model can be conducted by comparing the statistical metrics of the predicted tail quality by the calculation model with the difference in tail quality observed during the testing phase.

\subsubsection*{Fitting Quality and Generalization Performance}
Table~\ref{tab:verify_time} presents the fitting quality of inference time probability models obtained from training phase for all deep learning models across various servers and frameworks, as well as the generalization performance of the probability models in testing phase.
It is evident that all the rJSD values are below 0.05.
This indicates that the evaluation framework can effectively fit the probability distribution of inference times for each instance when estimating the inference time probability models for the influencing components.

Furthermore, each probability model demonstrates strong generalization on new data, indicating that the model comprehensively captures various scenarios during the deep learning inference process.
One reason for some test rJSD values being larger than their corresponding training phase values is that the testing phase involves fewer data points for fitting the probability density function.
Consequently, the fitting process during the testing phase lacks a significant amount of crucial information compared to the training phase, leading to discrepancies in the fitting results between the two phases.

\subsubsection*{Characterization of Tail Quality}
\input{tables/verfiy_quality}
By contrasting the differences in statistical indicators of these inference quality measurements, it is possible to preliminarily determine whether the tail quality calculation model possesses certain statistical characteristics.
The differences can be represented as $\varDelta = a - b$, where $a$ and $b$ is the worst-case values of training and testing stage, respectively.
Experiments employ tail latencies from 3 distinct percentiles - specifically, the 99th, 95th, and 90th percentiles of all inference times - as thresholds for computing tail quality.
Subsequently, the worst-case tail quality values are determined to observe the ability of prediction for tail quality in testing stage of the calculation model.
As illustrated in the Table~\ref{tab:verify_quality}, differences among all models on Server A are almost all below 0.
This suggests that the tail quality calculation model has achieved favorable estimations through the inference time probability model, and the value of the worst-case tail quality has been accurately predicted.

Furthermore, by examining the statistics of different tail quality in the Table~\ref{tab:verify_quality}, it is evident that the calculation models effectively capture the tail quality phenomena at various inference time thresholds.
As the thresholds become stricter, the worst-case tail quality gradually decreases.
This further validates the significance of tail quality, and the efficacy of this evaluation framework in predicting tail quality.

\subsubsection*{Efficiency of the Evaluation Framework}
\input{tables/verify_compute}
Due to the adoption of the heuristic algorithm for estimating the evaluation model, the computational resource expenditure in constructing this model depends on the number of iterations the algorithm takes to converge and the size of the dataset.
Table provides an account of the total inference count required for estimating the analysis model across all experiments.
This total inference count pertains to the fitting of PDF for inference times across all instances, rather than referring to inference rounds conducted on the entire dataset.
MLPerf Inference~\cite{reddiMLPerfInferenceBenchmark2020} stipulates that a total of 270,336 inferences should be conducted to statistically capture the 99th percentile tail latency.
As evident from the Table~\ref{tab:verify_compute}, the inference counts in all experiments are below this value except the model DETR-DC5.
The evaluation framework's average number of inference iterations across all models, compared to MLPerf, resulted in approximately 37.74\% reduction in computational workload.
Consequently, the computational resource expenditure of this evaluation framework is manageable in practical applications.

\subsection{Analysis of Influencing Components}
This section employs the instantiated framework to conduct a comprehensive analysis of the individual components that impact tail quality.
For the sake of conciseness and focused attention, this section primarily delves into the specific analysis of experiments related to the DETR-DC5 model in object detection tasks.

\begin{figure*}[!htbp]
\centering
\includegraphics[width=0.99\textwidth]{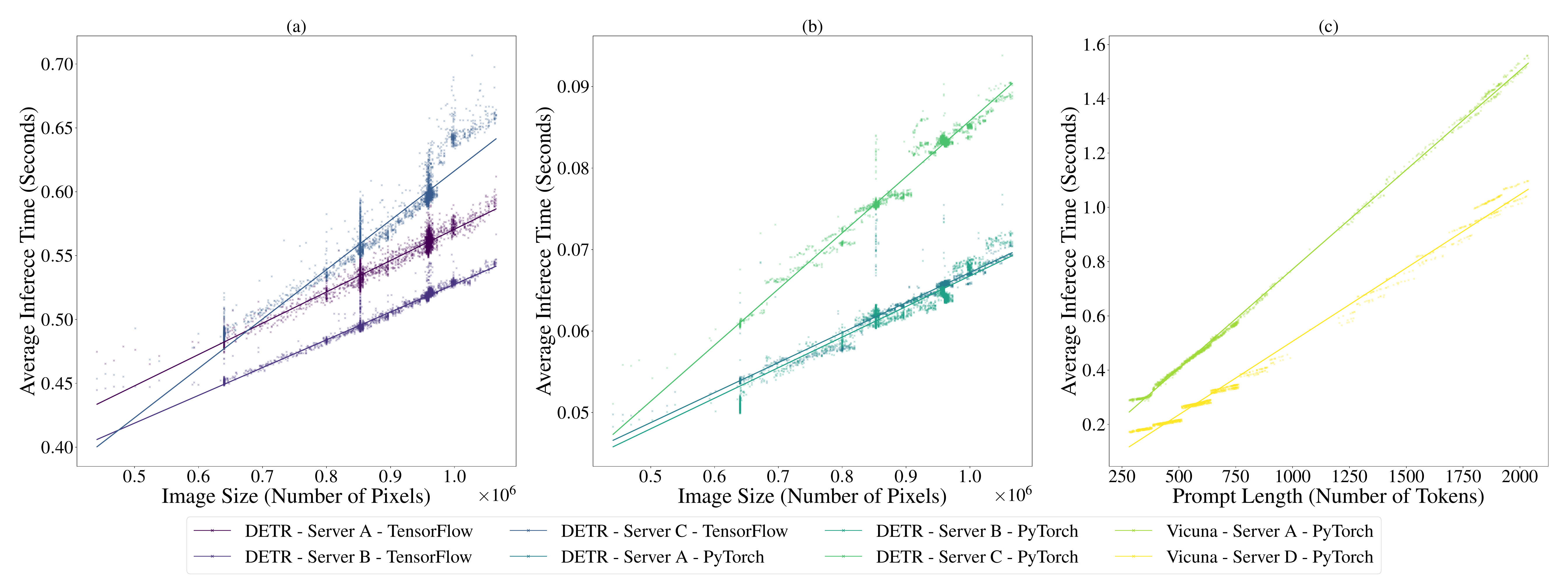}
\vskip -0.1in
\caption{
The relationship between the size of the instances of the dataset (COCO and MMLU) and the corresponding average inference time, which is established using linear regression.
Sub-figures (a) and (b) present experimental results for the DETR model on server A, B, and C.
Sub-figure (c) illustrates the experimental results for the Vicuna model implemented in the PyTorch framework on server A and D.
}
\label{fig:reg}
\end{figure*}
\subsubsection*{Effect of the Input Data}
\textbf{Impact of different instance sizes on inference time:}
Due to the characteristics of the DETR model and the nature of the object detection task, instances of the dataset isn't uniformly cropped to the same size before being fed into the model for inference.
Therefore, it is reasonable to assume that the inference time of the model may be influenced by the size of the input data, indicating a potential correlation between the two.
To validate this hypothesis, experiments conducted linear regression about the DETR model, examining the relationship between the number of pixels of input images and the time taken for inference on different deep learning frameworks and servers.
As depicted in Figure~\ref{fig:reg} (a) and (b), across all servers and regardless of whether TensorFlow or PyTorch is chosen as the deep learning framework, the model's inference time exhibits a positive linear correlation with the image size.

This confirms the hypothesis that as the image size increases, the computational workload of the deep learning system during inference also increases, subsequently leading to longer inference times.
Therefore, if the aim is to reduce model inference time, exploring solutions from the perspective of compressing image sizes could be a viable approach.
A similar phenomenon is observed in dialogue systems as well.
For chatbots like Vicuna, the inputs consist of sentences of varying lengths, also known as prompts, where the length is determined by the number of tokens after sentence segmentation.
In human-machine dialog scenarios, the input length during each model inference is highly likely to be different.
Therefore, as depicted in Figure~\ref{fig:reg} (c), the length of input sentences shows a positive linear correlation with the inference time of Vicuna.

\begin{figure}[!htbp]
\centering
\includegraphics[width=0.43\textwidth]{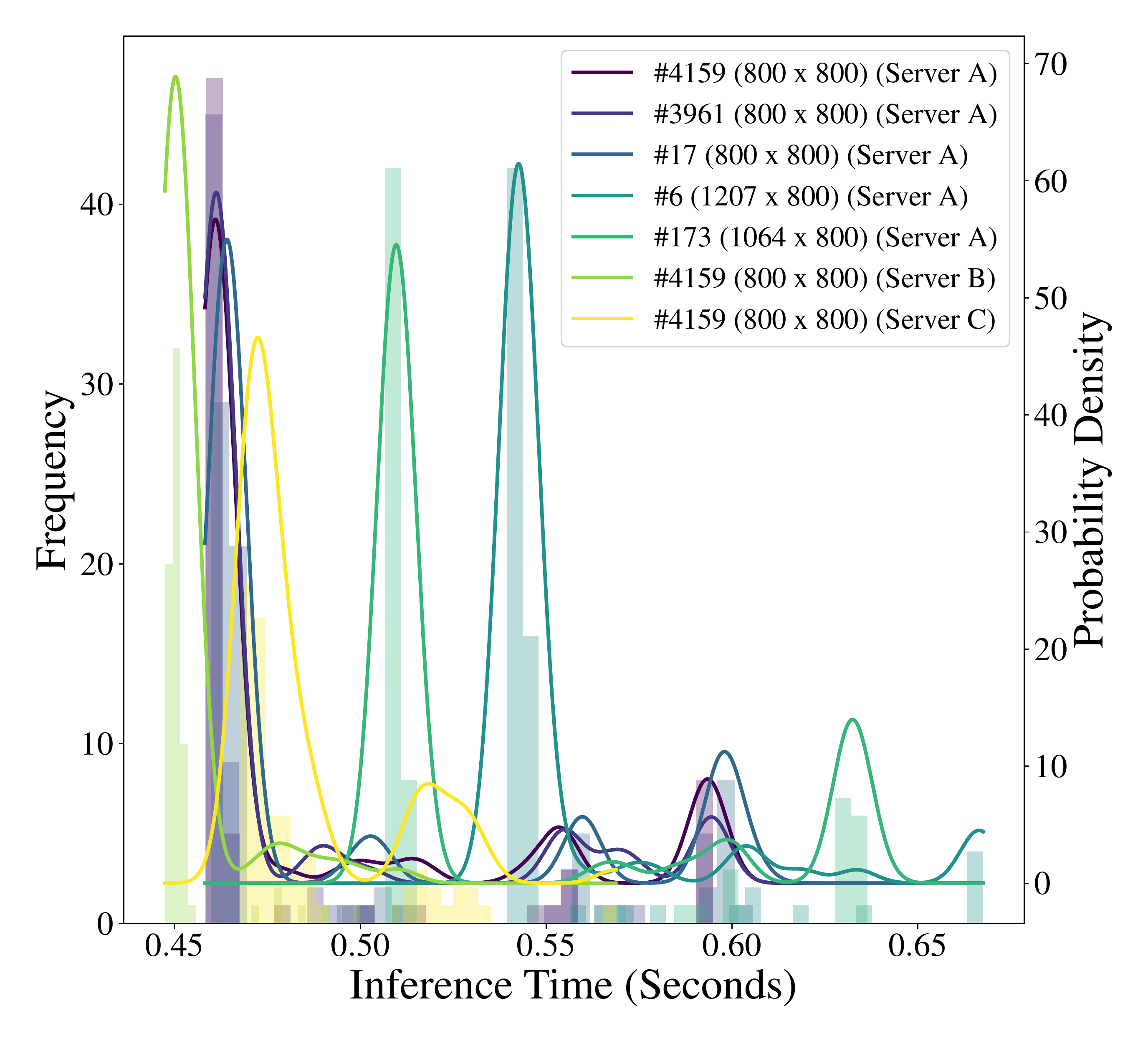}
\vskip -0.1in
\caption{
Fitting of probability density functions for inference times of instances with varying sizes, as well as for instances of the same size on different systems.
The histograms represent the frequencies of instance inference times corresponding to the left $y$ axis.
The values of the fitted probability density curves corresponding to the right $y$ axis.
}
\label{fig:fit}
\end{figure}

\noindent\textbf{Impact of same instance sizes on inference time:}
Interestingly, as evidenced by the scatter plot in Figure~\ref{fig:reg} (a) and (b), even when the image pixel count remains consistent and inferences are conducted on the same system, DETR model still exhibits significant fluctuations in inference time for different input samples.
To analyze the underlying reasons for these discrepancies, experiments compared the frequency distributions of inference times for different images of the same size.
As shown in Figure~\ref{fig:fit}, even though images $\#$4159, $\#$3961, and $\#$17 share the same dimensions, their frequency distributions for inference times still exhibit notable differences.
One possible reason for this phenomenon might be linked to the sparsity of images.
Furthermore, an insightful observation is that when categorizing images of the same size based on their width and height, it becomes apparent that images with longer heights tend to have relatively longer inference times.
Unfortunately, due to space constraints, the detailed analysis data is not presented here.
For a more comprehensive analysis, a finer-grained segmentation of the conditional random variables within the evaluation framework would be necessary.

\begin{figure}[!htbp]
\vskip -0.2in
\centering
\includegraphics[width=0.42\textwidth]{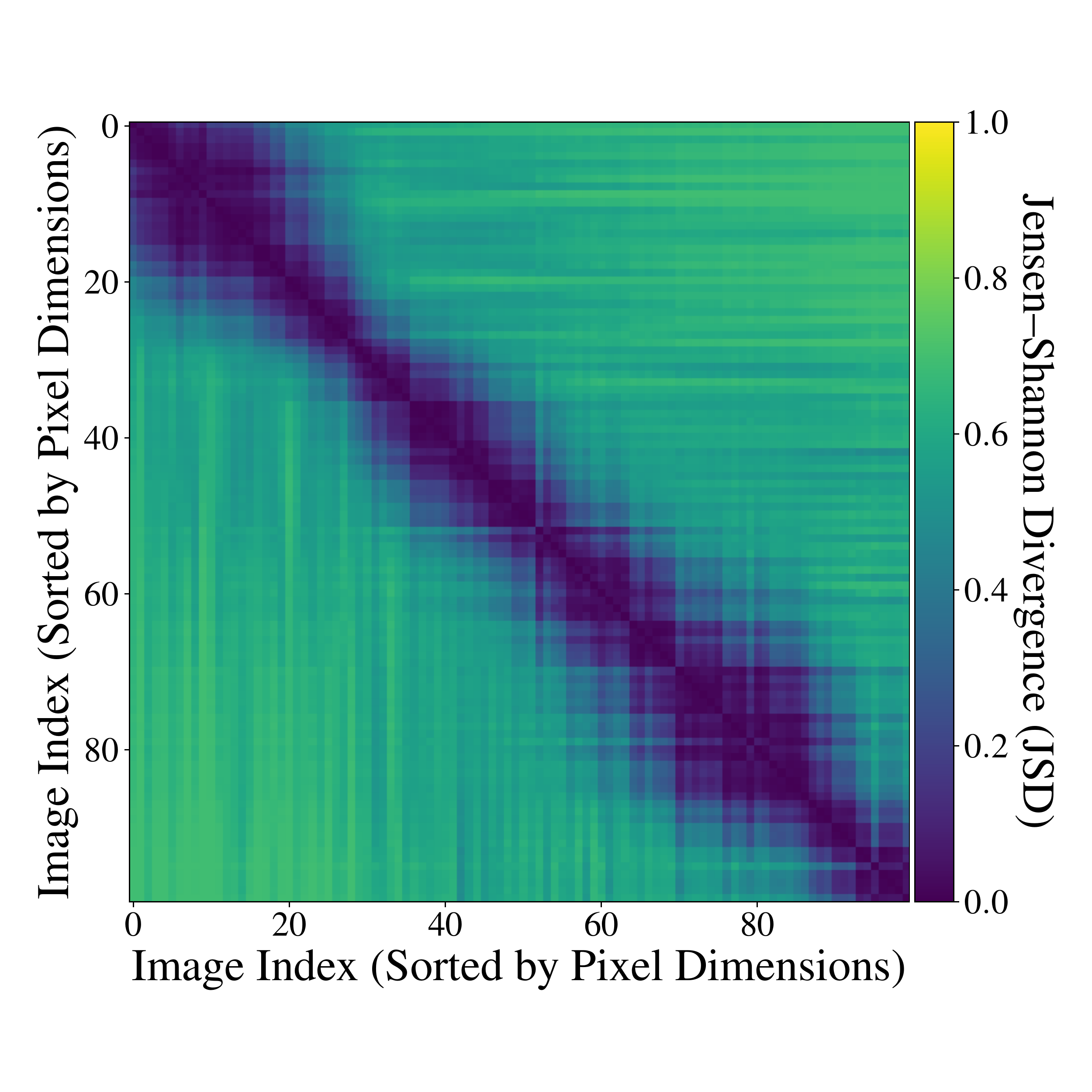}
\vskip -0.3in
\caption{
The Jensen-Shannon Divergence (JSD) between the probability density distributions of inference times fitted for each instance is depicted.
Both $x$ and $y$ axes of the heatmap are sorted in ascending order according to the size of the images.
This heatmap displays only 100 instances with distinct sizes present in the dataset.
}
\label{fig:heat}
\end{figure}

\noindent\textbf{Sensitivity of inference time analysis model to variations in input data:}
It's evident that input data indeed impact the model's inference time, though the differences in the effects generated by many instances are not excessively significant.
As inference time is positively correlated with input instance size, its distribution of statistical population should also exhibit a positive correlation with input size.
Figure~\ref{fig:heat} highlights that the evaluation framework's fitting of inference time distribution across different input data sizes nearly aligns with a positive correlation relationship.
Furthermore, as observed in Figure~\ref{fig:fit}, the evaluation framework also performs well in fitting models with very small distinctions in inference time frequency distribution.
Notably, the differences in the fitted curves corresponding to three images of the same size are clearly discernible.

\subsubsection*{Effect of the Hardware and Software Systems}
In addition to the analysis of input data, the instantiated evaluation framework also takes into account the influence of deep learning frameworks $\mathcal{F}$ and systems $\mathcal{S}$ on the inference results.
As the search space for $\mathcal{F}$ and $\mathcal{S}$ is significantly smaller than that of input data, this aspect of the analysis is relatively straightforward.
As granularity increases for the division of frameworks and systems, the complexity of analysis will increase along with the expansion of the search space.
However, it's worth noting that these considerations are beyond the scope of this work.

As evident from Figure~\ref{fig:reg} in the experiments with the DETR model, regardless of the framework used, Server C (Tesla V100) consistently exhibits shorter inference times compared to the other two servers.
Conversely, Server C (GeForce RTX 2080 Ti) notably performs worse than the other two servers, and its rate of increase in inference time with respect to increasing image size (slope of the regression line) is slower than that of the other servers.
This can also be observed from the fitting results of the inference time distribution for image $\#$4159 across different servers, as depicted in Figure~\ref{fig:fit}.
Also, the inference speed of PyTorch is notably higher than that of TensorFlow.
Furthermore, due to Vicuna's extensive parameter count of 13 billion, it necessitates the utilization of 4 GPUs for distributed model inference on Server A.
As depicted in the Figure~\ref{fig:reg}, the inference time on Server A remains considerably longer than that on Server C.
This can be attributed to a combination of factors, including GPU performance differences and the impact of data communication speed among GPUs during distributed inference.

%% file: tables/task_specs.tex
\begin{table}[!htbp]
\vskip -0.1in
\caption{
Task Specs.
The numbers placed in parentheses below the dataset names and model names represent the number of instances in dataset and the number of model parameters, respectively.
In these annotations, `K', `M', and `B' respectively denote `thousand', `million', and `billion'.
`mAP' and `Acc' respectively denote mean Average Precision and Accuracy.
}
\label{tab:task_specs}
\vskip 0.15in
\begin{center}
\begin{small}
\begin{tabular}{c|c|c|c|c}
\hline
\textbf{Area} & \textbf{Task}                                              & \textbf{Model} & \textbf{Data Set} & \textbf{Metric} \\ \hline\hline
CV        & \begin{tabular}[c]{@{}c@{}}Object\\ Detection\end{tabular} & \begin{tabular}[c]{@{}c@{}}DETR\\ ($\approx 60$M)\end{tabular}           & \begin{tabular}[c]{@{}c@{}}COCO\\ (5K)\end{tabular}              & mAP                     \\ \hline
CV   & \begin{tabular}[c]{@{}c@{}}Image\\ Classification\end{tabular} & \begin{tabular}[c]{@{}c@{}}ViT\\ ($\approx 306$M)\end{tabular}    & \begin{tabular}[c]{@{}c@{}}ImageNet\\ (50K)\end{tabular} & Acc \\ \hline
NLP & \begin{tabular}[c]{@{}c@{}}Dialogue\\ System\end{tabular}      & \begin{tabular}[c]{@{}c@{}}Vicuna\\ ($\approx 13$B)\end{tabular} & \begin{tabular}[c]{@{}c@{}}MMLU\\ (1531)\end{tabular}     & Acc \\ \hline
\end{tabular}
\end{small}
\end{center}
\vskip -0.1in
\end{table}

%% file: tables/verify_time.tex
\begin{table*}[tb]
\centering
\caption{
Square root of Jensen-Shannon divergence (rJSD) of all inference time analysis models corresponding to different deep learning models in the training and testing stage, with rJSD approaching 0 indicating that the fitting results are closer to the statistical population.
Note that all rJSD results in the table are the average values of the entire dataset.
N/A signifies the absence of relevant experiments.
}
\label{tab:verify_time}
\vskip 0.15in
\begin{small}
\begin{tabular}{c|c|c||cc||cc||cc||cc}
\hline
\multirow{2}{*}{\textbf{Model}} &
  \multirow{2}{*}{\textbf{Batch Size}} &
  \multirow{2}{*}{\textbf{Framework}} &
  \multicolumn{2}{c||}{\textbf{Server A}} &
  \multicolumn{2}{c||}{\textbf{Server B}} &
  \multicolumn{2}{c||}{\textbf{Server C}} &
  \multicolumn{2}{c}{\textbf{Server D}} \\ \cline{4-11} 
 &
   &
   &
  \multicolumn{1}{c|}{\textbf{Train}} &
  \textbf{Test} &
  \multicolumn{1}{c|}{\textbf{Train}} &
  \textbf{Test} &
  \multicolumn{1}{c|}{\textbf{Train}} &
  \textbf{Test} &
  \multicolumn{1}{c|}{\textbf{Train}} &
  \textbf{Test} \\ \hline\hline
\multirow{2}{*}{\begin{tabular}[c]{@{}c@{}}DETR\\ (DC5)\end{tabular}} &
  \multirow{2}{*}{1} &
  PyTorch &
  \multicolumn{1}{c|}{0.000} &
  0.004 &
  \multicolumn{1}{c|}{0.002} &
  0.036 &
  \multicolumn{1}{c|}{0.000} &
  0.007 &
  \multicolumn{1}{c|}{N/A} &
  N/A \\ \cline{3-11} 
 &
   &
  TensorFlow &
  \multicolumn{1}{c|}{0.032} &
  0.204 &
  \multicolumn{1}{c|}{0.025} &
  0.155 &
  \multicolumn{1}{c|}{0.010} &
  0.190 &
  \multicolumn{1}{c|}{N/A} &
  N/A \\ \hline
\multirow{2}{*}{DETR} &
  \multirow{2}{*}{2} &
  PyTorch &
  \multicolumn{1}{c|}{0.000} &
  0.004 &
  \multicolumn{1}{c|}{0.001} &
  0.011 &
  \multicolumn{1}{c|}{0.015} &
  0.207 &
  \multicolumn{1}{c|}{N/A} &
  N/A \\ \cline{3-11} 
 &
   &
  TensorFlow &
  \multicolumn{1}{c|}{0.030} &
  0.214 &
  \multicolumn{1}{c|}{0.018} &
  0.154 &
  \multicolumn{1}{c|}{0.006} &
  0.011 &
  \multicolumn{1}{c|}{N/A} &
  N/A \\ \hline
\multirow{2}{*}{ViT} &
  \multirow{2}{*}{256} &
  PyTorch &
  \multicolumn{1}{c|}{0.001} &
  0.011 &
  \multicolumn{1}{c|}{0.001} &
  0.013 &
  \multicolumn{1}{c|}{0.001} &
  0.015 &
  \multicolumn{1}{c|}{N/A} &
  N/A \\ \cline{3-11} 
 &
   &
  TensorFlow &
  \multicolumn{1}{c|}{0.011} &
  0.195 &
  \multicolumn{1}{c|}{0.006} &
  0.127 &
  \multicolumn{1}{c|}{0.009} &
  0.206 &
  \multicolumn{1}{c|}{N/A} &
  N/A \\ \hline
Vicuna &
  1 &
  PyTorch &
  \multicolumn{1}{c|}{0.006} &
  0.111 &
  \multicolumn{1}{c|}{N/A} &
  N/A &
  \multicolumn{1}{c|}{N/A} &
  N/A &
  \multicolumn{1}{c|}{0.000} &
  0.007 \\ \hline
\end{tabular}
\end{small}
\vskip -0.1in
\end{table*}

%% file: tables/verfiy_quality.tex
\begin{table*}[!htbp]
\centering
\caption{
The worst-case tail quality @$x$\%=$t$ of different models on Server A during both the training and testing phases.
A smaller $\varDelta$ indicates the better evaluation framework predictions for the worst-case tail quality.
}
\label{tab:verify_quality}
\vskip 0.15in
\begin{small}
\begin{tabular}{c|c||ccc||ccc||ccc||c}
\hline
 &
   &
  \multicolumn{3}{c||}{Tail Quality @$99$\%}  &
  \multicolumn{3}{c||}{Tail Quality @$95$\%}  &
  \multicolumn{3}{c||}{Tail Quality @$90$\%}  &
   \\ \cline{3-11}
\multirow{-2}{*}{\textbf{Model}} &
  \multirow{-2}{*}{\textbf{Framework}} &
  \multicolumn{1}{c|}{Train} &
  \multicolumn{1}{c|}{Test} &
  $\varDelta$ &
  \multicolumn{1}{c|}{Train} &
  \multicolumn{1}{c|}{Test} &
  $\varDelta$ &
  \multicolumn{1}{c|}{Train} &
  \multicolumn{1}{c|}{Test} &
  $\varDelta$ &
  \multirow{-2}{*}{Origin Quality} \\ \hline\hline
 &
  PyTorch &
  \multicolumn{1}{c|}{43.93} &
  \multicolumn{1}{c|}{44.29} &
  -0.36 &
  \multicolumn{1}{c|}{41.72} &
  \multicolumn{1}{c|}{42.35} &
  -0.62 &
  \multicolumn{1}{c|}{39.47} &
  \multicolumn{1}{c|}{40.03} &
  -0.56 &
  44.90 \\ \cline{2-12} 
\multirow{-2}{*}{\begin{tabular}[c]{@{}c@{}}DETR\\ (DC5)\end{tabular}} &
  TensorFlow &
  \multicolumn{1}{c|}{44.11} &
  \multicolumn{1}{c|}{44.18} &
  -0.07 &
  \multicolumn{1}{c|}{41.90} &
  \multicolumn{1}{c|}{42.20} &
  -0.30 &
  \multicolumn{1}{c|}{39.21} &
  \multicolumn{1}{c|}{39.45} &
  -0.24 &
  44.80 \\ \hline
 &
  PyTorch &
  \multicolumn{1}{c|}{43.03} &
  \multicolumn{1}{c|}{43.04} &
  -0.02 &
  \multicolumn{1}{c|}{41.10} &
  \multicolumn{1}{c|}{41.12} &
  -0.03 &
  \multicolumn{1}{c|}{39.00} &
  \multicolumn{1}{c|}{39.16} &
  -0.16 &
  43.50 \\ \cline{2-12} 
\multirow{-2}{*}{DETR} &
  TensorFlow &
  \multicolumn{1}{c|}{42.57} &
  \multicolumn{1}{c|}{42.79} &
  -0.22 &
  \multicolumn{1}{c|}{40.16} &
  \multicolumn{1}{c|}{40.32} &
  -0.16 &
  \multicolumn{1}{c|}{37.16} &
  \multicolumn{1}{c|}{37.69} &
  -0.54 &
  43.40 \\ \hline
 &
  PyTorch &
  \multicolumn{1}{c|}{78.53} &
  \multicolumn{1}{c|}{76.72} &
  \textbf{1.81} &
  \multicolumn{1}{c|}{65.36} &
  \multicolumn{1}{c|}{68.64} &
  -3.29 &
  \multicolumn{1}{c|}{48.00} &
  \multicolumn{1}{c|}{53.65} &
  -5.65 &
  81.51 \\ \cline{2-12} 
\multirow{-2}{*}{ViT} &
  TensorFlow &
  \multicolumn{1}{c|}{79.29} &
  \multicolumn{1}{c|}{79.73} &
  -0.44 &
  \multicolumn{1}{c|}{70.18} &
  \multicolumn{1}{c|}{72.31} &
  -2.13 &
  \multicolumn{1}{c|}{63.76} &
  \multicolumn{1}{c|}{65.62} &
  -1.87 &
  81.51 \\ \hline
LLM &
  PyTorch &
  \multicolumn{1}{c|}{51.99} &
  \multicolumn{1}{c|}{51.99} &
  0.00 &
  \multicolumn{1}{c|}{50.23} &
  \multicolumn{1}{c|}{50.16} &
  \textbf{0.07} &
  \multicolumn{1}{c|}{48.47} &
  \multicolumn{1}{c|}{48.47} &
  0.00 &
  52.71 \\ \hline
\end{tabular}
\end{small}
\vskip -0.1in
\end{table*}

%% file: tables/verify_compute.tex
\begin{table*}[!htbp]
\centering
\caption{
Total inference count of evaluation framework required for constructing the tail quality prediction model across various models and systems.
N/A signifies the absence of relevant experiments.
}
\label{tab:verify_compute}
\vskip 0.15in
\begin{small}
\begin{tabular}{cc||c||c||c||c||c}
\hline
\multicolumn{1}{c|}{\textbf{Model}}            & \textbf{Framework} & \textbf{Server A} & \textbf{Server B} & \textbf{Server C} & \textbf{Server D} & \textbf{MLPerf} \\ \hline\hline
\multicolumn{1}{c|}{DETR (DC5)} & PyTorch / TensorFlow            & 350,000           & 350000            & 350,000           & N/A               & 262,742         \\ \cline{1-7} 
\multicolumn{1}{c|}{DETR}     & PyTorch / TensorFlow            & 175,000           & 175000            & 175,000           & N/A               & 262,742         \\ \cline{1-7} 
\multicolumn{1}{c|}{ViT}      & PyTorch / TensorFlow            & 13,720            & 13720             & 13,720            & N/A               & 262,742         \\ \cline{1-7} 
\multicolumn{1}{c|}{LLM}                       & PyTorch / TensorFlow            & 107,170           & N/A               & N/A               & 107,170           & 262,742         \\ \hline
\multicolumn{2}{c||}{\begin{tabular}[c]{@{}c@{}}Average Inference Count\end{tabular}} &
  \begin{tabular}[c]{@{}c@{}}169,230\\ (64.41\%)\end{tabular} &
  \begin{tabular}[c]{@{}c@{}}179,573\\ (68.35\%)\end{tabular} &
  \begin{tabular}[c]{@{}c@{}}179,573\\ (68.35\%)\end{tabular} &
  \begin{tabular}[c]{@{}c@{}}107,170\\ (40.79\%)\end{tabular} &
  \begin{tabular}[c]{@{}c@{}}262,742\\ (100\%)\end{tabular} \\ \hline
\end{tabular}
\end{small}
\vskip -0.1in
\end{table*}

%% file: sections/related.tex
\section{Related Work}\label{related}
\subsection{Optimization of Inference Quality and Time}
\noindent\textbf{Domain generalization} aims to address the issue of significant inference quality degradation that may occur in well-trained deep learning models when facing unseen domains~\cite{wangGeneralizingUnseenDomains2021,zhouDomainGeneralizationSurvey2023}.
In contrast, our research primarily focuses on the phenomenon called tail quality, where the inference quality of deep learning models may fluctuate and experience a substantial decline when processing the same data.

\noindent\textbf{Efficient neural network inference} primarily focuses on increasing the computational efficiency of models under resource constraints, employing solutions like model quantization and model pruning, which often entail alteration of the original model~\cite{gongCompressingDeepConvolutional2014,jacobQuantizationTrainingNeural2018,guoEmpiricalStudyCharacterizing2019,gholamiSurveyQuantizationMethods2021}.
However, our work primarily concerns identifying inefficient inference processes and analyzing the underlying causes while keeping the model intact.

\subsection{Benchmarking of Deep Learning Inference}
\noindent\textbf{Evaluation of deep learning models} primarily focuses on specific inference quality metrics.
For instance, \emph{Average Precision (AP)} is the most popular metric that is used in various benchmark challenges for object detection, such as Pascal VOC~\cite{everinghamPascalVisualObject2010} and MS COCO~\cite{linMicrosoftCOCOCommon2014}.
\emph{Accuracy} and \emph{F-score} are utilized to evaluate most state-of-the-art classification models, such as large language models in the field of natural language processing~\cite{devlinBERTPretrainingDeep2019,zhengJudgingLLMasajudgeMTBench2023} and image classification models in autonomous driving, emotion recognition and healthcare~\cite{turayPerformingImageClassification2022,yurtseverSurveyAutonomousDriving2020,zhangDeepEmotionRecognition2021,rajpurkarDeepLearningChest2018,mckinneyAddendumInternationalEvaluation2020}.
Using only inference quality metrics can only reflect the optimal predictive ability of models on a specific test dataset.
Although, in addition to quality metrics, many studies also employ other system-related metrics to evaluate the processing speed of models~\cite{turayPerformingImageClassification2022}, such as \emph{floating-point operations per second (FLOPS)}~\cite{tanEfficientDetScalableEfficient2020,dosovitskiyImageWorth16x162023} and \emph{frames per second (FPS)}~\cite{liuSSDSingleShot2016}, the impact of changes in deep learning systems on inference quality and time are not taken into account.
These inference time metrics, such as FLOPS and FPS, can only reflect the average inference efficiency of the model on specific deep learning software and hardware systems.
When processing samples, they cannot illustrate how poorly models perform on different software and hardware systems.

\noindent\textbf{Benchmarking of deep learning systems} tends to prioritize inference time, throughput, and other system-related metrics.
MLPerf Inference~\cite{reddiMLPerfInferenceBenchmark2020} and AIBench~\cite{gaoAIBenchIndustryStandard2019,gaoAIBenchScalableComprehensive2019}, for example, utilizes \emph{tail latency} as its evaluation metric, while DAWNBench~\cite{colemanDAWNBenchEndtoEndDeep2017} adopts \emph{average inference latency}.
Although tail latency can reflect the stability and reliability of deep learning systems~\cite{deanTailScale2013,reddiMLPerfInferenceBenchmark2020}, helping identify issues for performance optimization, it cannot directly reflect the impact of variability and performance issues of systems on inference quality, as well as the potential adverse consequences may rise.
Indeed, these benchmark suites also incorporate inference quality as part of the evaluation procedure. Still, it is primarily used to set specific inference quality targets to ensure that the workloads meet the conditions imposed as benchmarks and is sufficient to assist in measuring the inference time consumed by different systems.

%% file: sections/conclusion.tex
\section{Conclusion}\label{conclusion}
The paper unveils a counterintuitive phenomenon that there are fluctuations in machine learning inference quality.
Authors coin a new term, ``tail quality,'' to characterize this phenomenon, overcoming existing evaluation methodologies limitations.
Due to the potential severe consequences of tail quality, such as loss of life or property damage, effective prediction and comprehensive analysis of tail quality are crucial.
This paper proposes a flexible and scalable evaluation framework, which can make reasonably accurate predictions of tail quality with lower computational costs than the state-of-the-practice like MLPerf Inference.
In conclusion, the authors aim to draw attention to ``tail quality'' phenomenon and call for exploration of evaluation methods based on the proposed evaluation framework in this paper.